\documentclass[journal]{IEEEtran}
\usepackage{amsmath,amsfonts}
\usepackage{algorithmic}
\usepackage{algorithm}
\usepackage{array}
\usepackage[caption=false,font=normalsize,labelfont=sf,textfont=sf]{subfig}
\usepackage{textcomp}
\usepackage{stfloats}
\usepackage{url}
\usepackage{verbatim}
\usepackage{graphicx}
\usepackage[sort,compress]{cite}
\usepackage{booktabs}
\usepackage{balance}
\usepackage{enumitem}
\usepackage{pifont}
\usepackage{listings}
\usepackage{xcolor}
\usepackage[compatibility=false]{caption}

\definecolor{keywordcyan}{rgb}{0.34, 0.78, 0.93}   
\definecolor{optionorange}{rgb}{0.95, 0.6, 0.13}  
\definecolor{framegray}{rgb}{0.88, 0.88, 0.88}      
\definecolor{promptbg}{HTML}{F8F9FA}      
\definecolor{promptframe}{HTML}{D1D5DB}   
\definecolor{promptkey}{HTML}{1D4ED8}     
\definecolor{prompttag}{HTML}{0F766E}     
\definecolor{promptcomment}{HTML}{6B7280} 

\lstdefinestyle{lightstyle}{
    backgroundcolor=\color{white},
    basicstyle=\color{black}\bfseries\ttfamily\scriptsize, 
    breaklines=true,
    frame=single,
    rulecolor=\color{framegray},
    captionpos=b,
    keepspaces=true,
    moredelim=[is][\bfseries\color{keywordcyan}]{**}{**},
    stringstyle=\color{optionorange},
    string=[b]',
}

\begin{document}

\title{Large Language Models Can Perform Automatic Modulation Classification via Discretized Self-supervised Candidate Retrieval}

\author{Mohammad~Rostami, Atik~Faysal, Reihaneh~Gh.~Roshan, Huaxia~Wang,~\IEEEmembership{Member,~IEEE,} Nikhil~Muralidhar,~\IEEEmembership{Member,~IEEE,} and Yu-Dong Yao,~\IEEEmembership{Fellow,~IEEE} 
\thanks{M. Rostami, A. Faysal, H. Wang are with the Department of Electrical and Computer Engineering, Rowan University, Glassboro, NJ, USA (e-mail: \{rostami23, faysal24, wanghu\}@rowan.edu).}
\thanks{R. Gh. Roshan and N. Muralidhar are with the Department of Computer Science (e-mail: \{rghasemi, nmurali1\}@stevens.edu), and Yu-Dong Yao is with the Department of Electrical and Computer Engineering (e-mail: yyao@stevens.edu). All authors are at the Stevens Institute of Technology, Hoboken, NJ, USA.}
\thanks{Code for this paper can be found at https://github.com/RU-SIT/DiSC-AMC}
}

\maketitle

\begin{abstract}
Identifying wireless modulation schemes is essential for cognitive radio, but standard supervised models often degrade under distribution shift, and training domain-specific wireless foundation models from scratch is computationally prohibitive. Large Language Models (LLMs) offer a promising training-free alternative via in-context learning, yet feeding raw floating-point signal statistics into LLMs overwhelms models with numerical noise and exhausts token budgets. We introduce DiSC-AMC, a framework that reformulates Automatic Modulation Classification (AMC) as an LLM reasoning task by combining aggressive feature discretization with nearest-neighbor retrieval over self-supervised embeddings. By mapping continuous features to coarse symbolic tokens, DiSC-AMC aligns abstract signal patterns with LLM reasoning capabilities and reduces prompt length by over $50$\%. Simultaneously, utilizing a DINOv2 visual encoder to retrieve the $k_\text{NN}$ most similar labeled exemplars provides highly relevant, query-specific context rather than generic class averages. On a 10-class benchmark, a fine-tuned 7B-parameter LLM using DiSC-AMC achieves $83.0$\% in-distribution accuracy ($-10$\,to\,$+10$\,dB) and $82.50$\% out-of-distribution (OOD) accuracy ($-11$\,to\,$-15$\,dB), outperforming supervised baselines.

Comprehensive ablations on vanilla LLMs demonstrate the token efficiency of DiSC-AMC. A training-free $7$B LLM achieves $71$\% accuracy using only $0.5$\,K-token prompt,surpassing a $200$B-parameter baseline that relies on a $2.9$K-token prompt. Furthermore, similarity-based exemplar retrieval outperforms naive class-average selection by over $20$\%. Finally, we identify a fundamental limitation of this pipeline. At extreme OOD noise levels ($-30$\,dB), the underlying self-supervised representations collapse, degrading retrieval quality and reducing classification to random chance.
\end{abstract}

\begin{IEEEkeywords}
Automatic modulation classification, large language models, prompt engineering, higher-order statistics.
\end{IEEEkeywords}

\section{Introduction}
\IEEEPARstart{A}{utomatic} Modulation Classification (AMC) enables cognitive radio systems to autonomously identify signal types for spectrum access and interference management \cite{jassim2022comparison}. While deep learning architectures such as Convolutional Neural Networks (CNNs) and Transformers achieve high accuracy in noisy environments \cite{peng2018modulation, faysal2024nmformer, faysal2025denomae}, they suffer from a fundamental rigidity: they are closed-set systems. These models fail when encountering signals outside their training distribution, requiring expensive data collection and retraining to adapt \cite{fontaine2024towards}. This lack of plasticity prevents deployment in dynamic, open-world wireless networks.

To address this, researchers are increasingly exploring Wireless Physical-layer Foundation Models (WPFMs) as a potential solution. However, training a WPFM from scratch is often impractical, requiring enormous computational power and massive datasets to combat the inherently noisy nature of wireless signals. To mitigate these high costs, our previous work introduced a plug-and-play (PnP) approach leveraging the In-Context Learning (ICL) capabilities of pre-trained Large Language Models (LLMs) \cite{rostami2025plug}. By treating signal statistics as text, LLMs can classify novel modulations from a few examples without the need for training new models from scratch.

Despite its potential, this approach currently faces a severe efficiency bottleneck. Prior methodologies directly inject raw, high-precision floating-point data into the prompt context. This strategy consumes a massive token budget, prohibiting real-time inference, and overwhelms the model with numerical noise rather than actionable patterns. Consequently, these implementations require prohibitively large models (e.g., 32B+ parameters) to achieve acceptable accuracy, rendering them unusable for resource-constrained edge devices.

To make foundation models viable for the wireless edge, we must bridge the gap between continuous signal physics and discrete language reasoning. We need a representation strategy that moves beyond raw numerical serialization to a compact, symbolic format that smaller, faster models can process effectively.

In this work, we introduce Discretized Self-supervised Candidate Retrieval Automatic Modulation Classification (DiSC-AMC), a token- and parameter-efficient framework for in-the-loop wireless reasoning. Our approach rests on the hypothesis that LLMs reason more effectively over abstract symbols than precise floating-point values. The DiSC-AMC pipeline achieves this via three key innovations: (i) a \textit{discretization mechanism} that maps any nummerical repressentation of signal including higher-order cumulants\cite{rostami2025plug} to compact symbolic tokens, reducing the input footprint; (ii) a \textit{dynamic context pruner} utilizing a lightweight DINOv2 visual encoder to select only the $k_\text{NN}$ nearest exemplars to the query as context; and (iii) a \textit{candidate selection} utilizing the same encoder to pick $k_\text{top}$ classes as the query's answer options, ensuring reliable predictions. This approach reduces token consumption by over 50\% and enables a lightweight $7$B-parameter model to achieve competitive accuracy, proving that efficient discretization is the key to practical wireless AI.

\begin{figure*}[!th]
\centering
\includegraphics[width=1\textwidth]{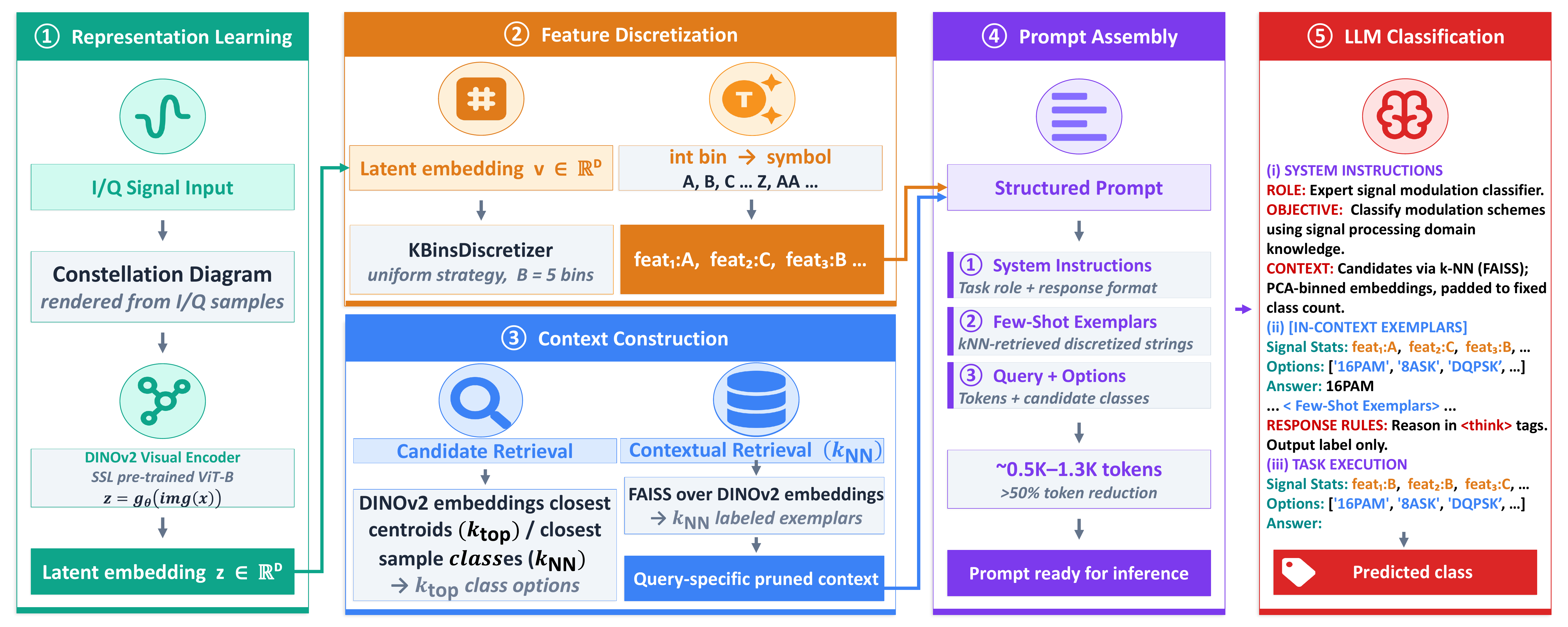}
\caption{
        \textbf{Overview of the DiSC-AMC Framework.}
        The pipeline comprises five stages:
        \textbf{(\ding{192}) Representation Learning}: 
            raw I/Q signals are rendered as constellation diagrams and encoded into latent embeddings
            $\mathbf{z} \in \mathbb{R}^{D}$ via a self-supervised DINOv2 visual encoder (ViT-B);
        \textbf{(\ding{193}) Feature Discretization}: 
            the continuous feature vector $\mathbf{v} \in \mathbb{R}^{D}$ is quantized using a
            \texttt{KBinsDiscretizer} into ordinal bins and mapped to compact alphabetic symbols
            (\texttt{A, B, C,~\ldots}), achieving $3$--$5\times$ token compression over raw
            floating-point serialization;
        \textbf{(\ding{194}) Context Construction}: 
            a two-stage retrieval module exploits DINOv2 embeddings to
            (a)~shortlist the $k_{\text{top}}$ most probable modulation classes via centroid proximity
            ($99.83\%$ recall at $k_{\text{top}}{=}5$) and
            (b)~retrieve the $k_{\text{NN}}$ most similar labeled exemplars via a FAISS index,
            forming a query-specific pruned context;
        \textbf{(\ding{195}) Prompt Assembly}: 
            the discretized query tokens, retrieved exemplars, and candidate class options are
            structured into a ${\sim}0.5$K--$1.3$K token prompt
            ($>50\%$ reduction relative to prior work~\cite{rostami2025plug}),
            comprising system instructions, few-shot exemplars, and the test query; and
        \textbf{(\ding{196}) LLM Classification}: 
            the structured prompt is fed to a QLoRA fine-tuned
            DeepSeek-R1-Distill-Qwen-7B model~\cite{guo2025deepseek},
            which performs chain-of-thought reasoning over the symbolic features
            before predicting the modulation class
            $\hat{y} \in \{\text{16QAM},\, \text{BPSK},\, \text{GFSK},\, \ldots\}$.
    }
\label{fig:pipeline}
\end{figure*}

\section{Related Work}

\subsection{Deep Learning in the Wireless Physical Layer}
AMC has progressed from likelihood-based hypothesis testing~\cite{dobre2007survey} and expert-crafted statistical features such as higher-order cumulants~\cite{mirarab2007robust} to data-driven, end-to-end learning architectures. CNNs~\cite{peng2018modulation, 8963964} and LSTMs~\cite{zhang2022deep} extract spatial and temporal correlations directly from raw I/Q data, while Transformer-based models capture long-range global dependencies via multi-head self-attention~\cite{faysal2024nmformer, ansari2025attention}. Despite achieving high accuracy in noisy environments, these supervised models operate as closed-set systems that degrade when exposed to out-of-distribution (OOD) data or novel channel conditions~\cite{fontaine2024towards}. Adapting them to new modulation schemes requires massive datasets and computationally expensive retraining.

\subsection{Foundations of In-Context Learning}
To overcome the rigidity of task-specific fine-tuning, In-Context Learning (ICL) has emerged as a transformative paradigm for LLMs~\cite{dong2024surveyincontextlearning}. In ICL, a frozen, pre-trained model is conditioned on a prompt containing a task description and a few demonstration examples, enabling accurate predictions for new queries without updating its internal parameters. Theoretical frameworks suggest that during the forward pass, Transformers dynamically simulate learning algorithms such as implicit gradient descent or Bayesian inference, allowing real-time task adaptation~\cite{dong2024surveyincontextlearning}. This training-free adaptability allows LLMs to incorporate human knowledge, switch between diverse tasks, and bypass the computational costs of continuous retraining. In complex classification scenarios, injecting Chain-of-Thought (CoT) reasoning narrows generative variability and bridges semantic gaps. Recent evidence further suggests that for implicit pattern detection tasks requiring deep logical rules, ICL can significantly outperform traditional fine-tuning by dynamically rewiring reasoning pathways rather than memorizing fixed abstractions~\cite{yin2024deeper}.

\subsection{In-Context Learning for Wireless Applications}
While highly successful in natural language processing, deploying ICL for wireless communications is an emerging research area that must bridge a significant semantic gap between continuous radio frequency (RF) physics and discrete language reasoning. The plug-and-play (PnP) framework~\cite{rostami2025plug} demonstrated that by treating signal statistics as text, LLMs can classify novel modulations from a few examples. However, directly serializing floating-point statistics exhausts strict token budgets and overwhelms models with numerical noise, degrading reasoning capabilities. More broadly, tokenization, the conversion of continuous data into discrete symbols, is a fundamental challenge whenever LLMs are applied to scientific domains. Research on symbolic representations of time series~\cite{10.1145/882082.882086} suggests that coarse-grained discretization can improve model performance by encouraging abstract pattern matching rather than overfitting to noisy numerical precision. This highlights the need for compact, symbolic representations that align with the LLM's discrete reasoning strengths, as well as dynamic retrieval mechanisms that provide query-specific context rather than generic exemplars.

Beyond the physical layer, ICL has shown potential in wireless network security and anomaly detection. LLMs such as GPT-4 have been applied to automatic network intrusion detection in wireless environments using illustrative, heuristic, and interactive ICL approaches. In the identification of distributed denial-of-service (DDoS) attacks, GPT-4 achieved over 95\% accuracy and a 90\% increase in F1-score using only 10 demonstration examples within the prompt, bypassing the need for costly fine-tuning~\cite{zhang2024large}.

\subsection{Exemplar Retrieval and Prompt Efficiency}
The performance of ICL is notoriously sensitive to the quality, quantity, and ordering of the demonstration examples~\cite{dong2024surveyincontextlearning, liu-etal-2022-makes}. Traditional $k$-NN selection retrieves semantically similar examples but incurs high computational overhead and fails to account for inter-exemplar interactions. Recent work formulates exemplar selection as a Multiple Exemplar-Subset Selection (MESS) problem: bandit-based frameworks model it as a stochastic linear bandit task, efficiently exploring optimal subsets while reducing expensive LLM evaluations~\cite{purohit2025sample}, and static subset scoring methods pre-select low-loss subsets that implicitly capture exemplar interactions~\cite{yang2024autoiclincontextlearninghuman}.

Beyond selection, example ordering drastically alters conditional token probabilities and final predictions, as LLMs often exhibit recency bias~\cite{bhope2025optiseq}. Adaptive ordering methods leverage the model's own log probabilities to evaluate permutations at inference time~\cite{bhope2025optiseq}, or filter orderings to ensure corpus-level label fairness while maximizing test-instance influence~\cite{guo2024demo}. ICL is also prone to label bias exacerbated by imbalanced demonstration sets, which calibration methods using content-free domain inputs can mitigate~\cite{dong2024surveyincontextlearning}. Despite these advances, ICL remains fundamentally constrained by context window size and the quadratic cost of processing lengthy demonstrations. Notably, most retrieval research focuses on optimizing \emph{which} exemplars to present; a complementary and largely unexplored strategy is to prune the \emph{label space} itself, reducing a $C$-way classification to a small multiple-choice set via a swappable shortlisting module, which can be implemented with any efficient mechanism such as a lightweight classifier, deterministic rules, or even another LLM. For real-time, low-latency applications at the wireless edge, balancing these computational costs with robust OOD generalization remains an open challenge.

\section{Methodology}
\label{sec:methodology}

We present \textbf{DiSC-AMC} (Discrete Signal Classification for Automatic Modulation Classification), a pipeline that converts raw in-phase/quadrature (I/Q) signals into compact symbolic prompts and leverages LLMs as reasoning-based classifiers. Our approach rests on the hypothesis that LLMs reason more effectively over abstract symbols than precise floating-point values. The pipeline comprises four stages, namely \emph{representation learning}, \emph{signal discretization}, \emph{dynamic context construction}, and \emph{LLM inference}, each described below.

\subsection{Problem Formulation}
\label{sec:problem}

Let $\mathbf{x} \in \mathbb{C}^{N}$ denote an observed baseband signal comprising $N$ complex I/Q samples received under an unknown modulation scheme $m \in \mathcal{M}$, where $\mathcal{M} = \{m_1, m_2, \dots, m_C\}$ is the set of $C$ candidate modulation classes. The goal of automatic modulation classification is to learn a mapping $f: \mathbb{C}^{N} \rightarrow \mathcal{M}$ that assigns $\mathbf{x}$ to its true class. In DiSC-AMC, $f$ is decomposed into a chain of transformations: $f = f_{\text{LLM}} \circ f_{\text{prompt}} \circ f_{\text{disc}} \circ f_{\text{feat}}$, where each component is detailed in the following subsections.

\subsection{Stage 1: Signal Representation and Feature Extraction}
\label{sec:features}

We support two parallel feature extraction pathways:

\paragraph{Statistical features.}
From the raw I/Q signal $\mathbf{x}$, we compute a statistical feature vector
\begin{equation}
\mathbf{s} = \bigl[\hat{\mu}_2, \dots, \hat{\mu}_{10},\; \hat{\kappa}_1, \dots, \hat{\kappa}_4,\; \hat{\gamma}_1, \hat{\gamma}_2,\; \hat{\sigma}_{\kappa_1}, \hat{\sigma}_{\kappa_2},\; \text{SNR}\bigr]^\top \in \mathbb{R}^{D_s},
\end{equation}
where $\hat{\mu}_k$ are sample central moments of order $k$, $\hat{\kappa}_j$ are $k$-statistics (the minimum-variance unbiased estimators of cumulants), $\hat{\gamma}_1$ and $\hat{\gamma}_2$ are skewness and kurtosis, and SNR is the signal-to-noise ratio. Higher-order cumulants are particularly discriminative: the fourth-order cumulant, for instance, achieves theoretical separation among BPSK, QPSK, and various QAM schemes~\cite{rostami2025plug}.

\paragraph{Encoder embeddings.}
Alternatively, the I/Q signal is rendered as a constellation diagram image and passed through a visual encoder $g_\theta$. The latent representation $\mathbf{z} = g_\theta(\text{img}(\mathbf{x})) \in \mathbb{R}^{D_z}$ is then reduced via PCA to $D_e \ll D_z$ components:
\begin{equation}
\mathbf{e} = \text{PCA}_{D_e}(\mathbf{z}) \in \mathbb{R}^{D_e}.
\end{equation}
This produces a compact, model-agnostic feature vector that captures learned visual patterns from the constellation geometry.

In both cases, the resulting feature vector (either $\mathbf{s}$ or $\mathbf{e}$) is standardized via a \texttt{StandardScaler} fitted on the training set.

\subsection{Stage 2: Discretization Mechanism}
\label{sec:discretization}

A central innovation of DiSC-AMC is the \emph{discretization mechanism} that maps any numerical signal representation to compact symbolic tokens, dramatically reducing the input footprint for the LLM. This approach is crucial because it normalizes feature scales and compels the model to focus on qualitative patterns rather than irrelevant decimal details.

Given a feature vector $\mathbf{v} \in \mathbb{R}^{D}$ (either $\mathbf{s}$ or $\mathbf{e}$), each feature dimension $v_j$ is independently quantized into one of $B$ ordinal bins via a \texttt{KBinsDiscretizer} with uniform strategy:
\begin{equation}
q_j = \left\lfloor \frac{v_j - v_j^{\min}}{v_j^{\max} - v_j^{\min}} \cdot B \right\rfloor, \quad q_j \in \{0, 1, \dots, B-1\}.
\end{equation}
The bin edges are fitted on training data and applied deterministically at test time. Each integer bin index $q_j$ is then encoded as a base-26 alphabetic symbol using the mapping
\begin{equation}
\phi(q_j) = \text{base26}(q_j) \in \{\texttt{A}, \texttt{B}, \dots, \texttt{Z}, \texttt{AA}, \dots\},
\end{equation}
where $\texttt{A}$ corresponds to the lowest bin and successive letters to higher bins. The full signal representation thus becomes a sequence of symbolic tokens:
\begin{equation}
\label{eq:disc}
f_{\text{disc}}(\mathbf{v}) = \bigl(\text{feat}_1\!:\;\phi(q_1),\; \text{feat}_2\!:\;\phi(q_2),\; \dots,\; \text{feat}_D\!:\;\phi(q_D)\bigr).
\end{equation}

This discretization offers three key benefits: (i)~it compresses each feature from a multi-digit floating-point string to one or two characters, reducing prompt token count by ${\sim}3\text{--}5\times$; (ii)~it abstracts away measurement noise, making the feature representation robust to small perturbations; and (iii)~ordinal letter codes align naturally with the symbolic reasoning capabilities of LLMs.

\subsection{Stage 3: Dynamic Context Construction}
\label{sec:context}

The effectiveness of ICL heavily depends on the quality and relevance of the provided examples. Using a large, fixed set of exemplars, as in prior work, is not only token-inefficient but can also introduce irrelevant information that degrades performance.

To address the token inefficiency and performance instability of ICL with large, static exemplar sets\cite{rostami2025plug}, we introduce a dynamic prompt pruning strategy. This method uses DINOv2~\cite{oquab2023dinov2} candidate retrieval to create a compact, query-specific context. For each signal, the encoder analyzes its constellation diagram to identify the $k_\text{top}$ most likely modulation classes (i.e., closest classes to the query or those with the highest softmax probabilities). The final prompt is then constructed using only the exemplars corresponding to this small, relevant subset, reframing the task as a constrained multiple-choice problem for the LLM.

We apply the candidate retrieval module to constellation diagram images of 10 modulation types across an SNR range of $-10$\,dB to $+10$\,dB. As shown in Fig.~\ref{fig:shortlist}, this candidate retrieval is highly effective; with $k_\text{top}=5$, it achieves 99.83\% accuracy, ensuring the correct class is almost always included in the candidate set provided to the LLM.

We construct the LLM prompt dynamically using two complementary mechanisms.

\subsubsection{Dynamic Context Pruner (Few-Shot Retrieval)}
\label{sec:rag}

Rather than providing a fixed set of few-shot examples, we employ a \emph{dynamic context pruner} that retrieves only the most relevant exemplars for each query signal. A FAISS~\cite{douze2024faiss} index $\mathcal{I}$ is built over the scaled feature vectors of all training signals. At inference time, for a query feature vector $\mathbf{v}_q$, we retrieve the $k_\text{NN}$ nearest training signals:
\begin{equation}
\mathcal{N}_{k_\text{NN}}(\mathbf{v}_q) = \operatorname*{arg\,top\text{-}k_\text{NN}}_{i \in \mathcal{D}_\text{train}} \; \frac{1}{\|\mathbf{v}_q - \mathbf{v}_i\|_2^2 + \epsilon},
\end{equation}
where $\epsilon > 0$ is a small constant for numerical stability. The retrieved exemplars are grouped by class label and added to the prompt as context, providing the LLM with signal-specific, diverse few-shot demonstrations.

To ensure class diversity, when a minimum-class constraint $c_{\min}$ is specified, the retrieval expands its search radius (up to $3k_\text{NN}$ candidates) and performs per-class brute-force scans to fill missing classes. This DINOv2 visual encoder ensures that the context window is populated with the most informative examples without exceeding the LLM's context budget.

\subsubsection{Candidate Retrieval}
\label{sec:candidate}

To further constrain the LLM's decision space, we employ a \emph{candidate retrieval} mechanism that narrows the full label set $\mathcal{M}$ down to the $k_\text{top}$ most plausible classes. We evaluate multiple candidate retrieval strategies, each utilizing the same DINOv2 visual encoder:

\begin{itemize}[leftmargin=*,nosep]
    \item \textbf{Centroid}: Euclidean distance from $\mathbf{z}$ to per-class prototype centroids $\boldsymbol{\mu}_c = \frac{1}{|\mathcal{D}_c|}\sum_{i \in \mathcal{D}_c} \mathbf{z}_i$; the $k_\text{top}$ nearest centroids are selected.
    \item \textbf{FAISS $k_\text{NN}$}~\cite{douze2024faiss}: Inverse-distance-weighted voting among $k_\text{NN}$-nearest neighbours in the FAISS index; the $k_\text{top}$ classes with the highest aggregate score are retained.
    \item \textbf{DNN head}: Softmax probabilities from a trained classification head; the $k_\text{top}$ highest-probability classes are selected.
    \item \textbf{Random Forest}: Class vote counts from an RF classifier trained on encoder features.
\end{itemize}

The selected candidate classes are presented to the LLM as a constrained multiple-choice option set, reducing the problem from $C$-way to $k_\text{top}$-way classification. This ensures reliable predictions by preventing the LLM from hallucinating out-of-vocabulary class labels.

\subsection{Stage 4: LLM Inference and Prompt Design}
\label{sec:llm}

The final prompt is assembled from three components:
\begin{enumerate}[nosep]
    \item \textbf{Instruction template}: A role description and response format guide, optionally enriched with source-aware context describing the candidate retrieval method and feature type.
    \item \textbf{Few-shot context}: The dynamically retrieved exemplars (Sec.~\ref{sec:rag}), each rendered as a discretized feature string paired with its ground-truth label.
    \item \textbf{Query}: The discretized feature string of the test signal, followed by the candidate class options (Sec.~\ref{sec:candidate}).
\end{enumerate}

The LLM is instructed to reason step-by-step using \texttt{<think>} tags before outputting its final classification, encouraging chain-of-thought deliberation over the symbolic features. An example of this structured prompt is illustrated in Fig.~\ref{fig:full_prompt}. 

\begin{figure}[h!]
\centering
\lstset{
    basicstyle=\ttfamily\scriptsize\color{black!90}, 
    backgroundcolor=\color{promptbg},
    rulecolor=\color{promptframe},
    frame=single,
    frameround=tttt,      
    breaklines=true,
    columns=fullflexible,
    escapechar=|,
    xleftmargin=0.5em,
    xrightmargin=0.5em,
    moredelim=[s][\color{promptkey}\bfseries]{**}{**},
    moredelim=[s][\color{prompttag}\bfseries]{<}{>},
    morecomment=[l][\color{promptcomment}\itshape]{...}
}

\begin{minipage}{0.95\linewidth}
\begin{lstlisting}
(i) **SYSTEM INSTRUCTIONS**
**ROLE:** Expert AI signal classifier for wireless modulation.
**OBJECTIVE:** Classify modulation schemes based on KBinsDiscretizer statistical features (moments, cumulants).
**CONTEXT:** Higher-order statistics (e.g., 4th-order cumulant) distinguish formats like BPSK/QAM.

(ii) **[IN-CONTEXT EXEMPLARS]**
**Signal Stats:** snr: E, skew: C, ... moment_{0..9}: [A..D], ... kstatvar_2: A
**Options:** ['16PAM', '8ASK', 'DQPSK', '4ASK', '4PAM']
**Answer:** 16PAM
.
.
... (k_top pruned examples) ...

**RESPONSE RULES:**
1. MUST use <think> tags to reason about features vs. options.
2. Output ONLY the classification label (e.g., 16PAM).

(iii) **TASK EXECUTION**
**Signal Stats:** snr: A, skew: C, kurt: B, ... kstatvar_2: A
**Options:** ['16PAM', '8ASK', 'DQPSK', '4ASK', '4PAM']
**Answer:**
\end{lstlisting}
\end{minipage}

\caption{A condensed representation of the DiSC-AMC prompt structure. The prompt includes (i) system instructions defining the task and feature set, (ii) few-shot in-context exemplars mapping discretized statistics to modulation labels, and (iii) the final query with constrained decoding rules.}
\label{fig:full_prompt}
\end{figure}

\section{Experimental Setup}
\subsection{Dataset}
\paragraph{Synthetic dataset.}
Following the evaluation protocol of the PnP framework~\cite{rostami2025plug}, we generate a controlled synthetic dataset using an identical signal generation procedure. The dataset comprises I/Q signals from $C=10$ digital modulation classes: 4ASK, 4PAM, 8ASK, 16PAM, CPFSK, DQPSK, GFSK, GMSK, OOK, and OQPSK. We generate 10,000 training samples and a balanced evaluation set of 200 query signals across a range of -10 dB to +10 dB. Additionally, we generate 200 query signals for out-of-distribution (OOD) evaluation across a range of -15 dB to -11 dB.
\paragraph{RadioML.2018.01a~\cite{8267032}.}
To evaluate OOD generalization over unseen modulation classes and channel conditions, we test on RadioML.2018.01a~\cite{8267032}, a 24-class benchmark spanning digital and analog modulation schemes with realistic channel impairments (Rician fading, AWGN, frequency and phase drift). This dataset is entirely disjoint from the synthetic training set. We evaluate at five representative SNR levels ($-20$, $-10$, $0$, $+10$, $+20$\,dB), using 2{,}400 training and 240 test samples per SNR level (12{,}000 training and 1{,}200 test samples in total).
\subsection{Baselines}
To rigorously evaluate DiSC-AMC, we benchmark against prior methods across two distinct settings: training-free in-context learning and supervised fine-tuning.
\paragraph{Training-Free LLM Baselines} For training-free inference, our primary baseline is the PnP framework~\cite{rostami2025plug}. Unlike our discretized and retrieval-augmented approach, PnP directly prompts models using raw floating-point statistical features alongside a comprehensive, unpruned set of exemplars. To isolate the performance gains of our representation and retrieval mechanisms, we apply both PnP and DiSC-AMC to powerful open-weight reasoning models, specifically DeepSeek-R1-Distill-Qwen-7B and DeepSeek-R1-Distill-Qwen-32B~\cite{guo2025deepseek}.
\paragraph{Supervised and OOD Baselines} To contextualize our fine-tuned LLM's performance against specialized wireless models, we compare it against DenoMAE2.0~\cite{faysal2025denomae2}, a leading denoising masked autoencoder for modulation classification. We also include results from its predecessor, DenoMAE~\cite{faysal2025denomae}, and NMformer~\cite{faysal2024nmformer}, a Transformer-based architecture built specifically for noisy signals. This comparison establishes whether an LLM reasoning over discretized features can compete with architectures structurally engineered for the wireless physical layer.
\paragraph{OOD Encoder Baselines vs.\ DiSC-AMC} To evaluate OOD generalization, we compare standalone encoder classification heads against our full DiSC-AMC pipeline on RadioML.2018.01a at SNR levels $-20$\,dB to $+20$\,dB under two encoder settings: (i)~a \emph{fine-tuned} setting where the encoder backbone is updated on the target dataset before adding a classification head (Table~\ref{tab:OOD_finetune}), and (ii)~a \emph{frozen} setting where the encoder weights remain fixed and only the classification head is trained (Table~\ref{tab:OOD_frozen}). This assesses whether pairing an encoder backbone with DiSC-AMC's discretized retrieval-augmented reasoning yields stronger OOD performance than the encoder's own supervised head, regardless of whether the encoder has been adapted to the target domain.

\paragraph{DiSC-AMC Configuration (Training-Free, Gemini API)}
We evaluate our three-stage pipeline using Google's Gemini models~\cite{comanici2025gemini} accessed via their public API, with discretized statistical tokens (bins\,=\,5), $k_\text{top}$\,=\,5 candidate retrieval, and fixed exemplar:
\begin{itemize}
    \item {\textit{Gemini-2.5-Flash~\cite{comanici2025gemini}:}} A highly efficient model optimized for speed and low-cost inference.
    \item {\textit{Gemini-2.5-Pro~\cite{comanici2025gemini}:}} A state-of-the-art, high-performance model.
\end{itemize}

These models were selected for their diverse positions on the performance-efficiency spectrum and their accessibility via a free public API, which facilitates reproducible research. This configuration serves as the testbed for \emph{prompt engineering ablations} (Sec.~\ref{sec:ablations_trainingfree}), isolating the effects of discretization granularity, candidate set size ($k_\text{top}$), and prompt format on classification accuracy and token budget.

\paragraph{DiSC-AMC Configuration (Fine-tuned, Local LLMs)}
DiSC-AMC couples discretized statistical tokens (5 bins) with FAISS $k_\text{NN}$ exemplar retrieval ($k_\text{NN}{=}50$) over DINOv2 embeddings, topped by $k_\text{top}{=}5$ DINOv2 candidate retrieval. We deploy two open-weight LLMs locally:
\begin{itemize}
    \item \textit{DeepSeek-R1-Distill-Qwen-7B~\cite{guo2025deepseek}}: a 7B-parameter reasoning-distilled model.
    \item \textit{GLM-4.6V-Flash~\cite{hong2025glm}}: a 4.6B-parameter vision--language model.
\end{itemize}

This configuration serves as the testbed for \emph{pipeline architecture ablations} (Sec.~\ref{sec:ablations_finetuned}), isolating four design axes: (a)~retrieval strategy (FAISS vs.\ centroid); (b)~feature representation (self-supervised embeddings vs.\ discretized statistics); (c)~RAG-augmented vs.\ fixed exemplar selection; and (d)~in-distribution vs.\ OOD robustness. Classification accuracy on 200 test queries is the primary metric.

\section{Experimental Results}
Table~\ref{tab:main_results} presents the in-distribution comparison. Among supervised baselines, DenoMAE2.0~\cite{faysal2025denomae2} leads with 82.40\% accuracy. Remarkably, our DiSC-AMC framework, using a 7B open-weight LLM with FAISS-retrieved contextual exemplars encoded via DINOv2 self-supervised embeddings, achieves \textbf{83.00\%}, surpassing all supervised baselines. Table~\ref{tab:ood_main} further shows that DiSC-AMC maintains \textbf{82.50\%} OOD accuracy, far exceeding the encoder-only baselines.

Tables~\ref{tab:OOD_finetune} and~\ref{tab:OOD_frozen} present OOD generalization on RadioML.2018.01a (24 classes). Under the fine-tuned encoder setting (Table~\ref{tab:OOD_finetune}), DINOv2-based DiSC-AMC reaches 76.25\% at $+20$\,dB compared to just 9.58\% for the fine-tuned encoder alone, an $8{\times}$ improvement. Under the frozen encoder setting (Table~\ref{tab:OOD_frozen}), DiSC-AMC with DINOv2 achieves 45.42\% at $+20$\,dB versus 7.08\% for the frozen encoder head, demonstrating that DiSC-AMC's retrieval-augmented reasoning amplifies encoder backbones even without any target-domain fine-tuning. In both settings, standalone encoders plateau below 15\% across all SNR levels, confirming that the encoder's classification head alone cannot generalize to unseen modulation classes, while DiSC-AMC's structured prompting enables meaningful OOD performance.

Table~\ref{tab:efficiency} highlights the token and parameter inefficiency of brute-force prompting. When using raw 2.9K-token prompts, accuracy reaches merely 5.20\% on a 7B model and 47.80\% on a 32B model, with even the 200B o3-mini achieving only 69.92\%. In contrast, DiSC-AMC leverages compact, contextually retrieved prompts to unlock the reasoning capabilities of much smaller models. This demonstrates that intelligent prompt construction is more critical than sheer model scale. Having established DiSC-AMC's competitiveness across settings, we next dissect the pipeline to identify which design choices and prompt engineering decisions drive this performance.

\begin{table}[!th]
\centering
\caption{In-Distribution Supervised Accuracy}
\label{tab:main_results}
\begin{tabular}{lcc}
\hline
\textbf{Model}                                    & \textbf{Params} & \textbf{Acc.\ (\%)} \\
\hline
DINOv2                                           & 86M                  & 65.33              \\
Nmformer\cite{faysal2024nmformer}                & 86M                  & 71.60              \\
ViT\cite{faysal2025denomae2}                     & 86M                  & 79.90              \\
DEiT\cite{faysal2025denomae2}                    & 86M                  & 81.20              \\
MoCov3\cite{faysal2025denomae2}                  & 86M                  & 81.00              \\
BEiT\cite{faysal2025denomae2}                    & 86M                  & 80.40              \\
MAE\cite{faysal2025denomae2}                     & 86M                  & 80.10              \\
DenoMAE\cite{faysal2025denomae}                  & 86M                  & 81.30              \\
DenoMAE2.0\cite{faysal2025denomae2}              & 86M                  & 82.40              \\
\textbf{DiSC-AMC (Ours)}                          & 7B   & \textbf{83.00} \\
\hline
\end{tabular}
\caption*{\textit{DeepSeek-R1-Distill-Qwen-7B}, 5 bins, FAISS $10_\text{NN}$ exemplar retrieval and candidate retrieval, using DINOv2 embeddings as features.}
\end{table}

\begin{table}[!th]
\centering
\caption{OOD Accuracy: Encoder Baselines vs.\ DiSC-AMC}
\label{tab:ood_main}
\begin{tabular}{lcc}
\hline
\textbf{Model}            & \textbf{Params} & \textbf{OOD Acc.\ (\%)} \\
\hline
DINOv2                    & 86M  & 63.90  \\
DenoMAE2.0\cite{faysal2025denomae2} & 86M  & 63.25  \\
\textbf{DiSC-AMC (Ours)}  & 7B   & \textbf{82.50} \\
\hline
\end{tabular}
\caption*{Evaluated on the synthetic OOD set. DiSC-AMC: \textit{DeepSeek-R1-Distill-Qwen-7B}, 5 bins, FAISS $10_\text{NN}$ exemplar retrieval and candidate retrieval, DINOv2 embeddings.}
\end{table}

\begin{table}[!th]
\centering
\caption{OOD Generalization: Fine-tuned Encoder vs.\ DiSC-AMC}
\label{tab:OOD_finetune}
\resizebox{\columnwidth}{!}{%
\begin{tabular}{ccccc}
\hline
 & \multicolumn{2}{c}{\textbf{Finetune Encoder (\%)}} & \multicolumn{2}{c}{\textbf{DiSC-AMC (\%)}} \\
\cline{2-3} \cline{4-5}
\textbf{SNR (dB)} & \textbf{DINOv2} & \textbf{DenoMAE2.0} & \textbf{DINOv2} & \textbf{DenoMAE2.0} \\
\hline
$-20$ & 3.33  & 4.17  & \textbf{19.58} & 17.92 \\
$-10$ & 3.33  & 5.42  & 17.50 & \textbf{19.17} \\
$0$   & 6.67  & 8.75  & \textbf{40.83} & 25.00 \\
$+10$ & 8.75  & 13.33 & \textbf{72.50} & 24.17 \\
$+20$ & 9.58  & 11.25 & \textbf{76.25} & 21.25 \\
\hline
\end{tabular}
}
\caption*{Evaluated on RadioML.2018.01a (24 classes). ``Finetune Encoder'' columns use a supervised classification head with finetuned encoder. DiSC-AMC uses each finetuned encoder as the backbone within our pipeline (\textit{DeepSeek-R1-Distill-Qwen-7B}, bins\,=\,5, $k_\text{NN}$\,=\,10, $k_\text{top}$\,=\,5).}
\end{table}

\begin{table}[!th]
\centering
\caption{OOD Generalization: Frozen Encoder vs.\ DiSC-AMC}
\label{tab:OOD_frozen}
\resizebox{\columnwidth}{!}{%
\begin{tabular}{ccccc}
\hline
 & \multicolumn{2}{c}{\textbf{Encoder Only (\%)}} & \multicolumn{2}{c}{\textbf{DiSC-AMC (\%)}} \\
\cline{2-3} \cline{4-5}
\textbf{SNR (dB)} & \textbf{DINOv2} & \textbf{DenoMAE2.0} & \textbf{DINOv2} & \textbf{DenoMAE2.0} \\
\hline
$-20$ & 3.75  & 3.33  & 10.83 & \textbf{15.42} \\
$-10$ & 5.00  & 2.92  & \textbf{11.67} & 5.83 \\
$0$   & 6.67  & 9.58  & \textbf{28.33} & 12.5 \\
$+10$ & 15.00  & 14.58 & \textbf{36.25} & 5.83 \\
$+20$ & 7.08  & 12.92 & \textbf{45.42} & 6.25 \\
\hline
\end{tabular}
}
\caption*{Evaluated on RadioML.2018.01a (24 classes). ``Encoder Only'' columns use a supervised classification head while the encoder is frozen. DiSC-AMC uses each encoder as the backbone within our pipeline (\textit{DeepSeek-R1-Distill-Qwen-7B}, bins\,=\,5, $k_\text{NN}$\,=\,10, $k_\text{top}$\,=\,5).}
\end{table}

\begin{table}[!th]
\centering
\caption{Token and Parameter Efficiency: PnP Baseline vs.\ DiSC-AMC}
\label{tab:efficiency}
\resizebox{\columnwidth}{!}{
\begin{tabular}{lcccc}
\hline
\textbf{Model}                                   & \textbf{Parameters}  & \textbf{\# Tokens} & \textbf{Accuracy (\%)} \\
\hline
\textit{DeepSeek-R1}\cite{rostami2025plug}   & 7B                   & 2.9K               & 05.20                \\
\textit{DeepSeek-R1}\cite{rostami2025plug}   & 32B                  & 2.9K               & 47.80                \\
\textit{o3-mini}\cite{rostami2025plug}           & 200B                 & 2.9K               & 69.92                \\
\textbf{\textit{DeepSeek-R1 (DiSC-AMC)}}                  & 7B                   & \textbf{0.5K}               & \textbf{71.9}        \\
\hline 
\end{tabular}
} 
\caption*{\textit{DeepSeek-R1} refers to \textit{DeepSeek-R1-Distill-Qwen}. DiSC-AMC row: bins\,=\,5, FAISS $10_\text{NN}$ exemplar retrieval and candidate retrieval, using DINOv2 embeddings as features.}
\end{table}

\section{Ablation Studies}
\label{sec:ablations}

DiSC-AMC introduces three interacting mechanisms (discretization, retrieval-augmented exemplar selection, and candidate pruning) each with tunable design choices. To understand their individual and joint contributions, we organize our ablations into two complementary tracks. Section~\ref{sec:ablations_finetuned} examines \emph{pipeline architecture} decisions: which retrieval strategy, feature representation, and exemplar selection method to use, evaluated on locally deployed open-weight LLMs. Section~\ref{sec:ablations_trainingfree} then isolates \emph{prompt engineering} choices, how discretization granularity, candidate set size, and prompt format affect token efficiency and accuracy, evaluated via the Gemini API. Together, these two tracks decompose the end-to-end pipeline into interpretable components and justify the default configuration used in the main results.

\subsection{Pipeline Architecture Ablations}
\label{sec:ablations_finetuned}

\begin{figure}[!th]
\centering
\includegraphics[width=\linewidth]{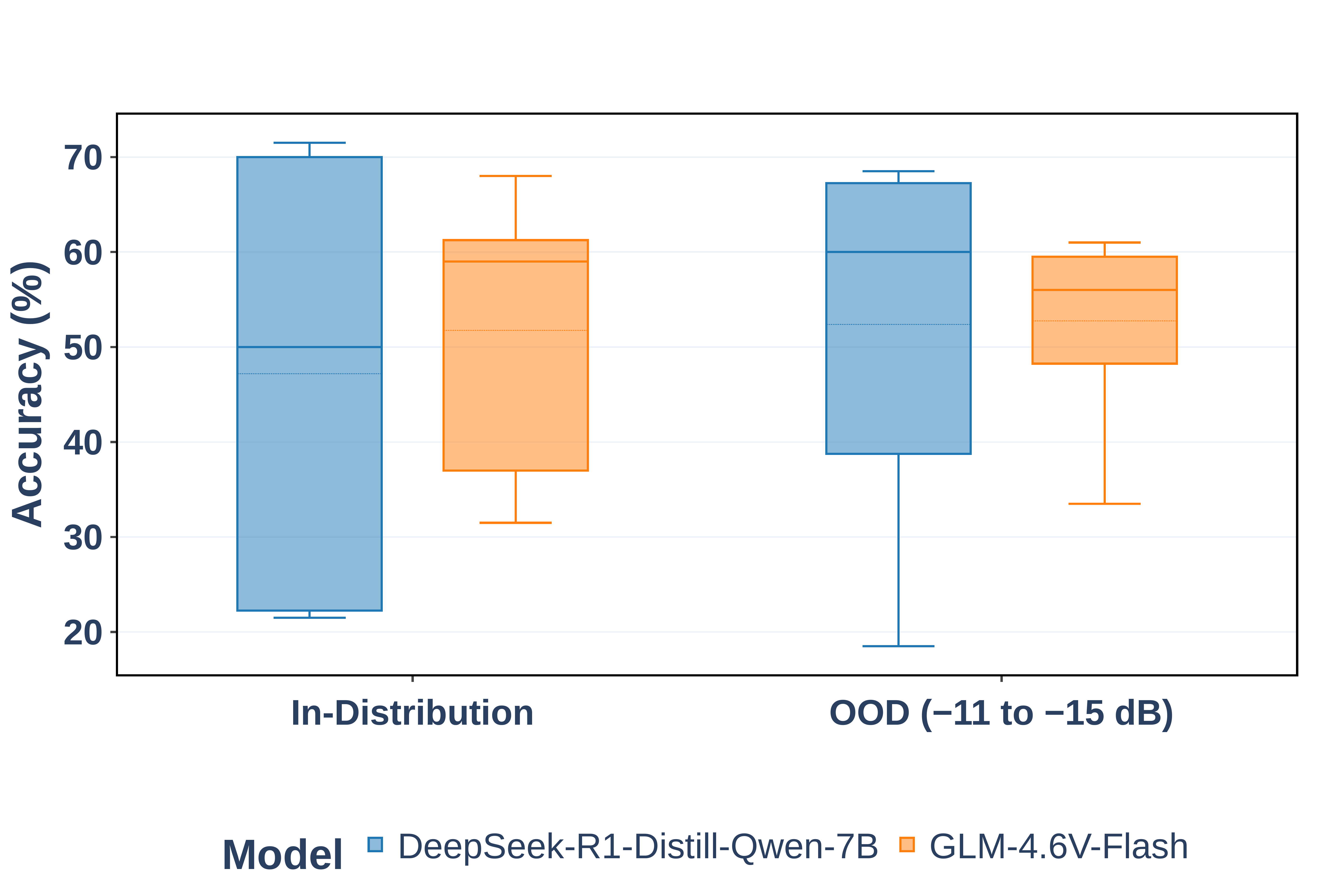}
\caption{LLM classification accuracy across in-distribution and OOD environments, by model (DeepSeek-R1-Distill-Qwen-7B and GLM-4.6V-Flash). Boxes show the spread over all retrieval configurations.}
\label{fig:dist_overview}
\end{figure}

Using \textit{DeepSeek-R1-Distill-Qwen-7B} and \textit{GLM-4.6V-Flash} in a training-free setting, we vary four independent design axes: (1)~candidate retrieval strategy, (2)~feature representation, (3)~contextual exemplar retrieval (RAG), and (4)~distribution shift. Fig.~\ref{fig:dist_overview} provides an overview of per-model accuracy across all three evaluation environments.

\subsubsection{Candidate Retrieval Strategy: FAISS vs.\ Centroid}
\begin{figure}[!th]
\centering
\includegraphics[width=\linewidth]{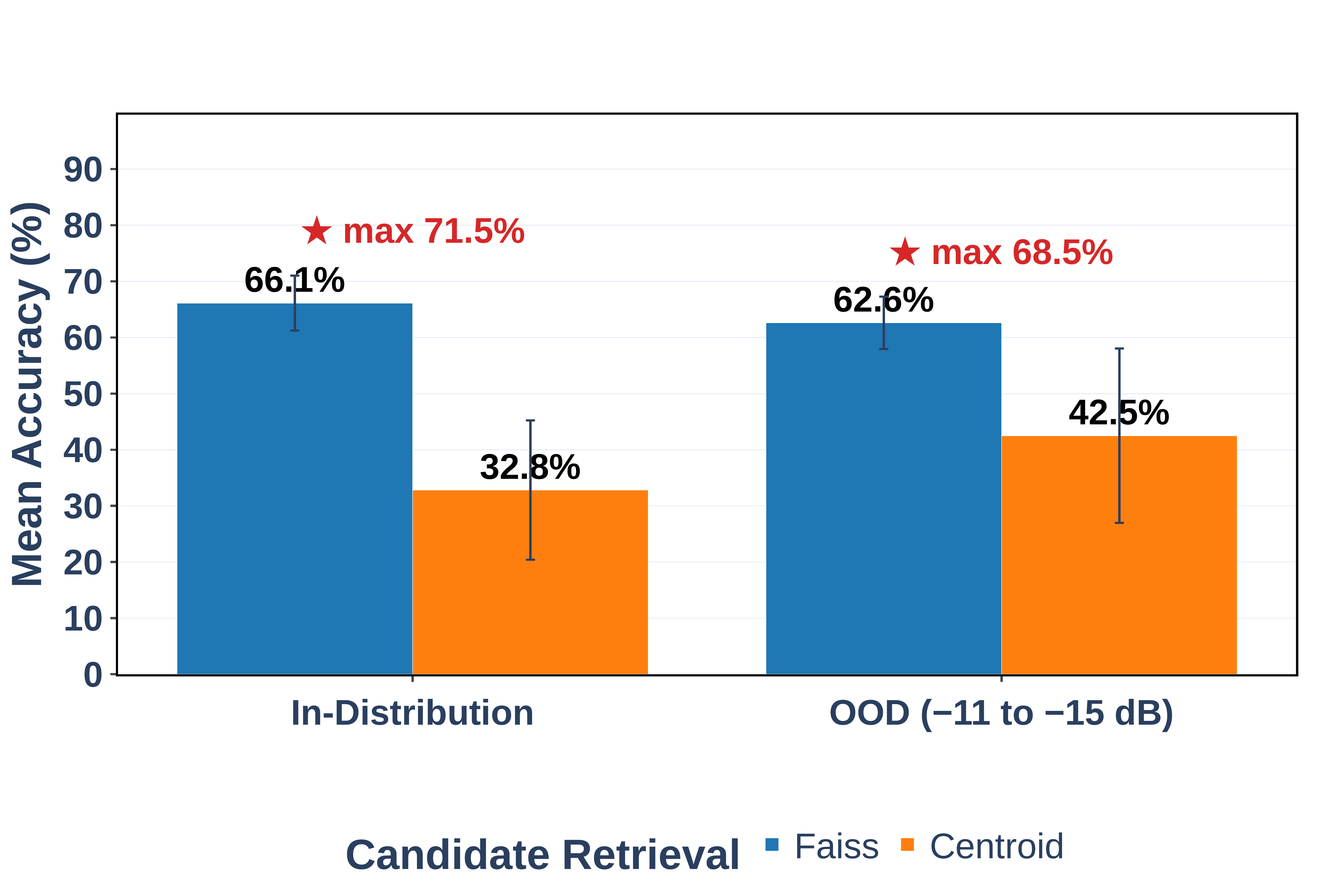}
\caption{Candidate retrieval strategy comparison: FAISS $10_\text{NN}$ vs.\ centroid selection, grouped by distribution environment. Error bars show standard deviation across all model and feature configurations.}
\label{fig:filter_strategy}
\end{figure}

As shown in Fig.~\ref{fig:filter_strategy}, FAISS $k_\text{NN}$ retrieval substantially outperforms centroid-based exemplar selection. In-distribution, FAISS achieves a mean accuracy of $66.1\%$ versus $32.8\%$ for centroid; under moderate OOD shift ($-11$ to $-15$\,dB), FAISS maintains $62.6\%$ versus $42.5\%$ for centroid. Centroid-selected exemplars cluster around class means and lack the query-specific diversity required for effective in-context learning, whereas FAISS retrieves nearest neighbors that are more informative for the query at hand.

A key factor underlying this gap is \emph{SNR matching}: because signal statistics (moments, cumulants) vary substantially with SNR, exemplars drawn from a different noise level carry a mismatched statistical signature that confuses rather than guides the LLM's reasoning. FAISS retrieval naturally tends to preserve SNR proximity, whereas centroid selection does not.

\subsubsection{Feature Representation: Embeddings vs.\ Statistics}
\begin{figure}[!th]
\centering
\includegraphics[width=\linewidth]{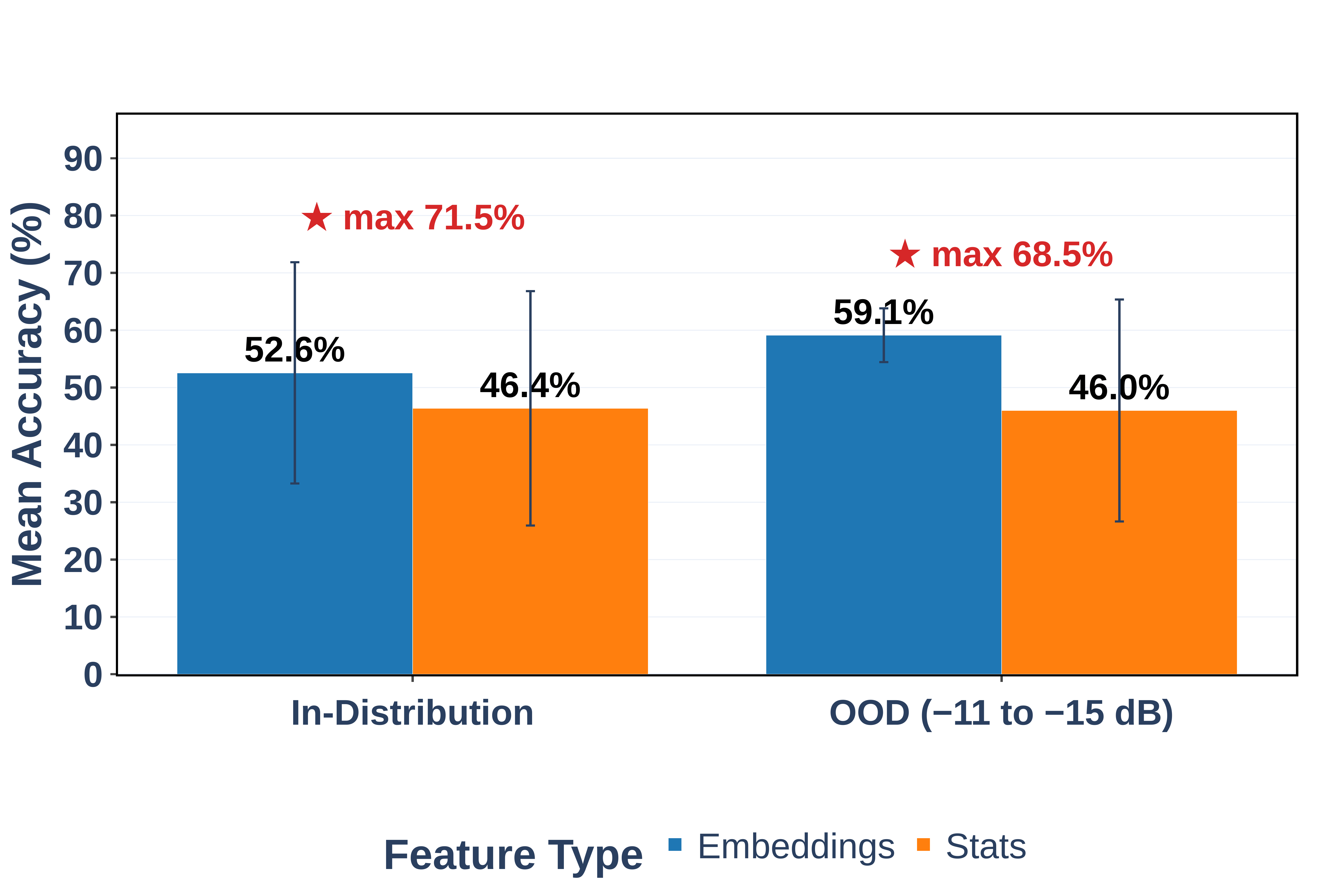}
\caption{Feature representation comparison: DINOv2 embeddings vs.\ discretized statistics as FAISS retrieval features, grouped by distribution environment.}
\label{fig:feature_comparison}
\end{figure}

Replacing discretized statistics with DINOv2 embeddings as the retrieval feature consistently improves accuracy (Fig.~\ref{fig:feature_comparison}). The best embedding-based configuration (DeepSeek-7B + FAISS Candidate Retreival + embeddings) reaches $71.5\%$ in-distribution, versus $69.0\%$ for the best statistics-only configuration. The self-supervised encoder captures visual invariances of constellation diagrams that cumulant features do not, providing a richer similarity signal for FAISS retrieval.

\subsubsection{Impact of Contextual Retrieval (RAG)}
\begin{figure}[!th]
\centering
\includegraphics[width=\linewidth]{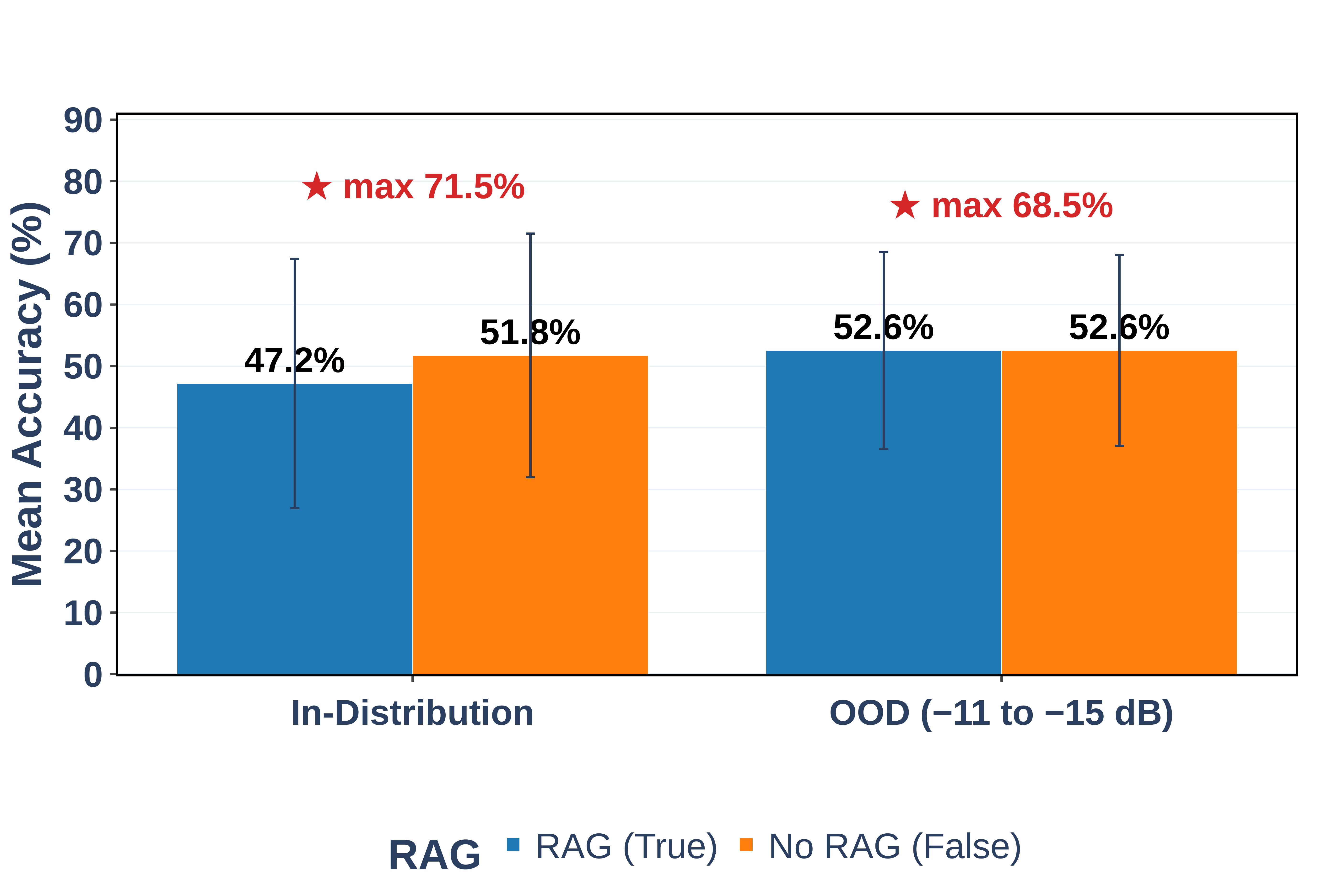}
\caption{Impact of contextual retrieval (RAG): dynamic FAISS-retrieved exemplars vs.\ fixed exemplar selection, grouped by distribution environment. Error bars show standard deviation across all model and feature configurations.}
\label{fig:rag_impact}
\end{figure}

RAG-augmented and fixed-exemplar pipelines achieve comparable mean accuracies in-distribution ($47.2\%$ vs.\ $51.8\%$, averaged across all retrieval and feature configurations), as shown in Fig.~\ref{fig:rag_impact}. However, RAG's dynamic retrieval of relevant exemplars catches up to the fixed set's performance at $52.6\%$

\subsubsection{Out-of-Distribution Generalization}
Accuracy degrades gracefully under moderate OOD shift. The best FAISS configuration retains 68.5\% at $-11$ to $-15$\,dB, only 3 percentage points below its in-distribution score of 71.5\%. Under extreme shift ($-30$\,dB), however, all configurations collapse to near-random levels (${\sim}$5--10\%). At this noise level, constellation diagrams become effectively featureless; the DINOv2 encoder's embeddings lose discriminative structure, degrading both retrieval quality and LLM classification to random chance. This is a fundamental limitation of retrieval-based pipelines that rely on visual signal representations.

\subsection{Prompt Engineering Ablations}
\label{sec:ablations_trainingfree}

Table~\ref{tab:ablation_disc} isolates the effect of prompt format in the training-free setting (centroid candidate retrieval, RAG\,=\,false). The 7B DeepSeek model under the PnP prompt achieves only 9\%, confirming that raw floating-point serialization overwhelms small models. In contrast, our 5B \textit{Gemini-2.5-Flash} with a 1.3K-token DiSC-AMC prompt reaches 45.5\%, comparable to the 32B DeepSeek PnP baseline (32.5\%) at less than half the token cost. With \textit{Gemini-2.5-Pro} at 0.9K tokens, accuracy reaches 51.0\%, the best among all training-free configurations. All DiSC-AMC results substantially exceed the $\frac{1}{10}$ random-chance baseline, confirming that the LLM is reasoning over the provided context rather than hallucinating.

The following ablations use \textit{Gemini-2.5-Flash} to isolate individual prompt design choices. As a reference point consistent with the SNR-matching finding in Sec.~\ref{sec:ablations_finetuned}, centroid-based exemplar selection yields only 8.63\% and random selection 16.47\% under this configuration, confirming that exemplar quality governs performance regardless of model family.

\subsubsection{Effect of Prompt Size ($k_\text{top}$)}
\begin{figure}[!th]
\centering
\includegraphics[width=\linewidth]{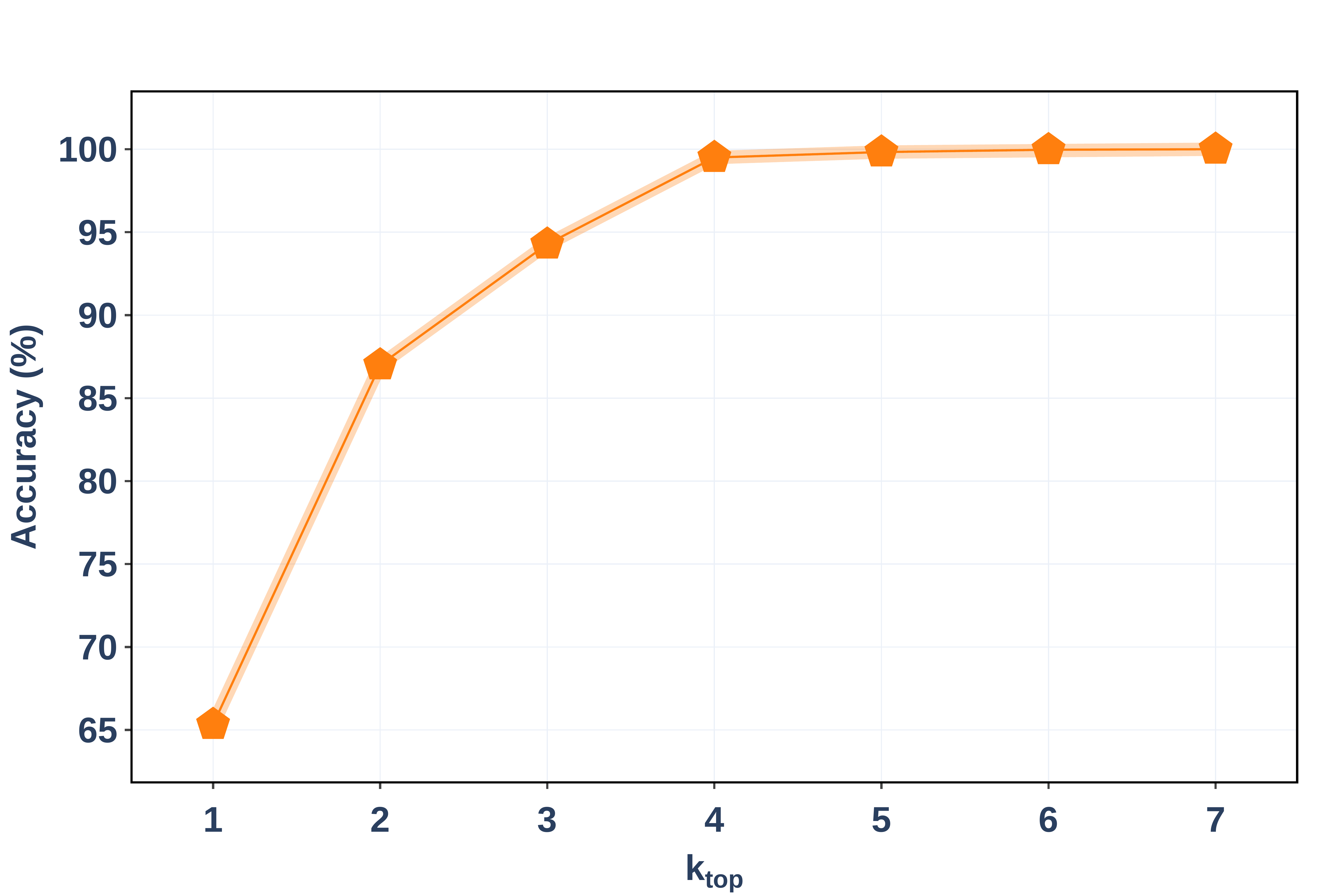}
\caption{DINOv2 candidate retrieval accuracy vs.\ $k_\text{top}$. At $5_\text{top}$ the correct class is included 99.83\% of the time.}
\label{fig:shortlist}
\end{figure}

\begin{figure}[!th]
\centering
\includegraphics[width=\linewidth]{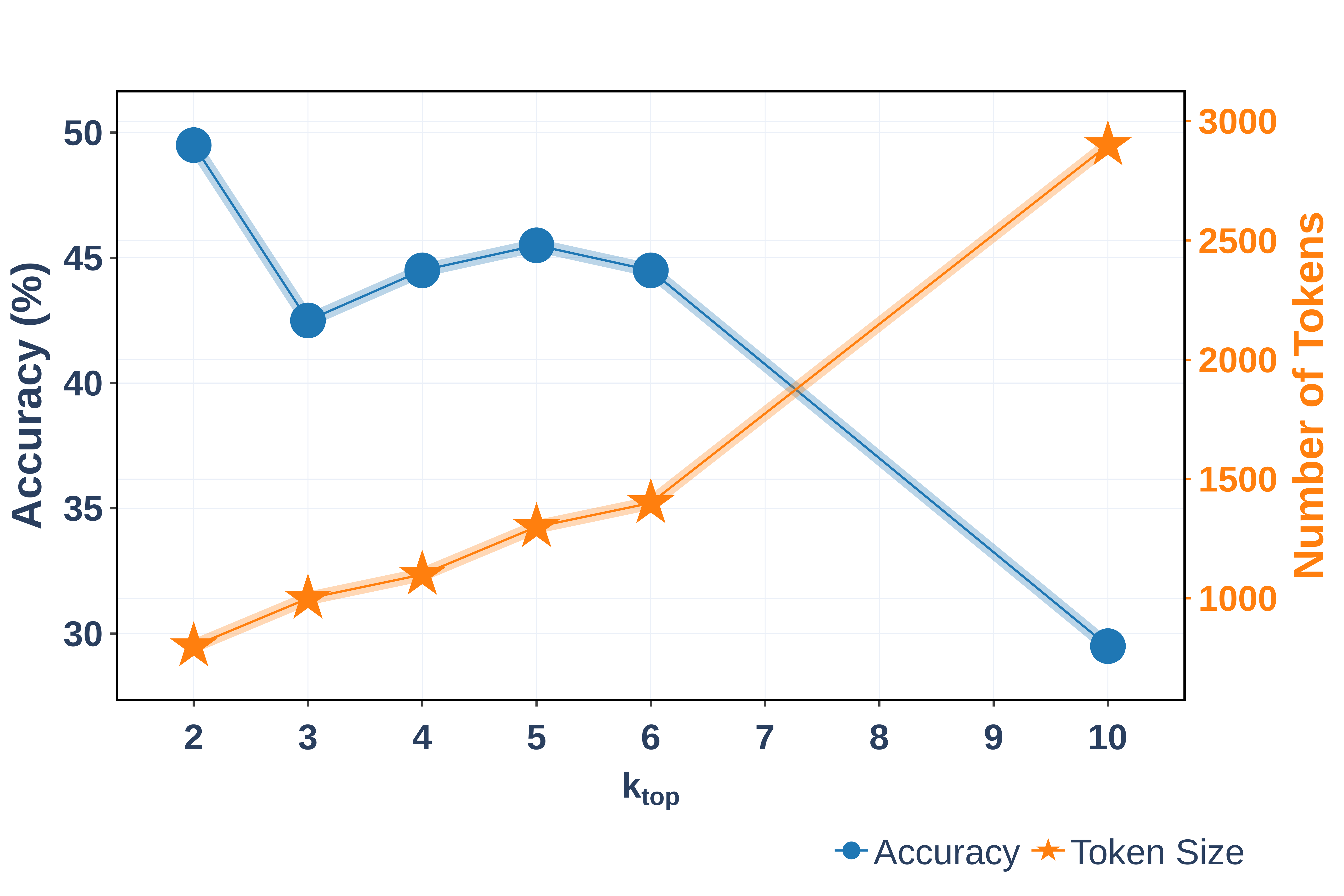}
\caption{Effect of $k_\text{top}$ on accuracy (left axis) and prompt token count (right axis) (\textit{Gemini-2.5-Flash}, 5 bins). Classification accuracy peaks at $5_\text{top}$ while token cost grows monotonically.}
\label{fig:ktops}
\end{figure}

Fig.~\ref{fig:ktops} shows accuracy and token count as a function of $k_\text{top}$ (5 bins). Increasing $k_\text{top}$ from 4 to 5 yields a marginal accuracy gain (44.5\%$\to$45.5\%) at the cost of 0.1K additional tokens, a diminishing return. At $10_\text{top}$ with 10 bins, the prompt expands to 2.9K tokens and accuracy drops sharply to 29.5\%. The candidate retrieval accuracy of the DINOv2 shortlisting module reaches 99.83\% at $5_\text{top}$ (Fig.~\ref{fig:shortlist}), making further increases unnecessary. Taken together, these results support a ``less is more'' principle: a concise, focused context outperforms a large one containing redundant or distracting exemplars.

\subsubsection{Effect of Discretization Granularity}
\begin{figure}[!th]
\centering
\includegraphics[width=\linewidth]{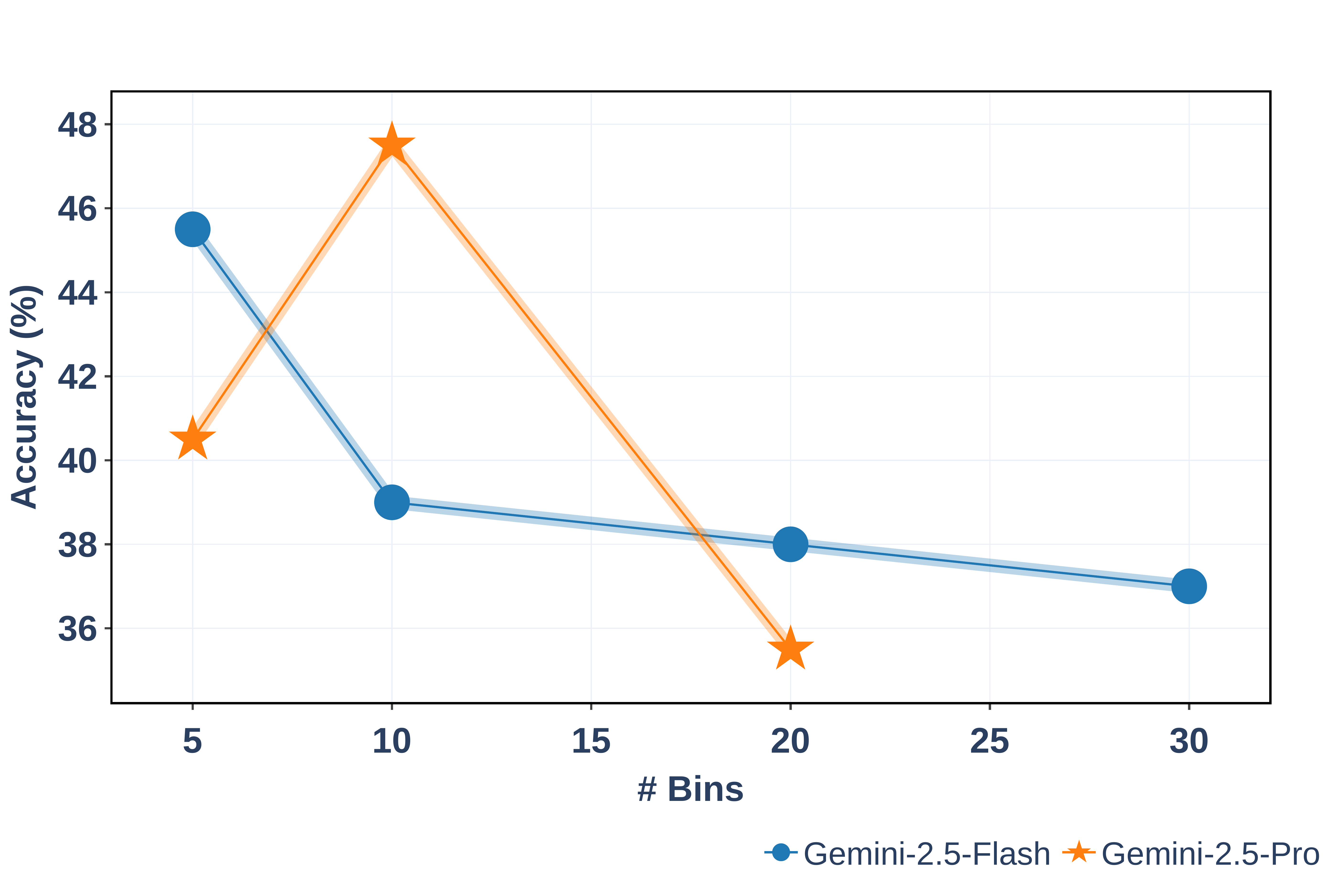}
\caption{Effect of discretization granularity on accuracy ($5_\text{top}$). \textit{Gemini-2.5-Flash} (blue) peaks at 5 bins; \textit{Gemini-2.5-Pro} (orange) peaks at 10 bins.}
\label{fig:bins}
\end{figure}

Fig.~\ref{fig:bins} shows that the optimal number of bins is model-dependent. \textit{Gemini-2.5-Flash} performance peaks at 45.5\% with 5 bins and degrades monotonically with finer granularity. \textit{Gemini-2.5-Pro}'s accuracy is non-monotonic, peaking at 47.5\% with 10 bins before declining. The more capable Pro model can leverage slightly more feature detail, but for both models excessively fine-grained discretization hurts performance. This indicates that coarse symbolic codes align better with LLM reasoning over noisy physical-layer signals than high-precision numerical representations.

\begin{table}[!th]
\centering
\caption{Effect of Prompt Format on Training-Free LLMs (Centroid candidate retrieval, RAG\,=\,false)}
\label{tab:ablation_disc}
\resizebox{\columnwidth}{!}{%
\begin{tabular}{lcccc}
\hline
\textbf{Model}                                   & \textbf{Parameters}  & \textbf{\# Tokens} & \textbf{Accuracy (\%)} \\
\hline
\textit{DeepSeek-R1}\cite{rostami2025plug}   & 7B                   & 2.9K               & 05.20                \\
\textit{DeepSeek-R1}\cite{rostami2025plug}   & 32B                  & 2.9K               & 47.80                \\
\textit{o3-mini}\cite{rostami2025plug}           & 200B                 & 2.9K               & 69.92                \\
\hline
\textit{DeepSeek-R1}                         & 7B                   & 2.9K               & 09.00                   \\
\textit{DeepSeek-R1}                         & 32B                  & 2.9K               & 32.50                   \\
\textit{Gemini-2.5-Flash}                        & 5B                   & 2.9K               & 29.50                 \\
\textit{Gemini-2.5-Pro}                          & -                    & 2.9K               & 42.50                   \\
\hline
\textit{DeepSeek-R1} (ours)                      & 7B                   & 1.3K               & \textbf{33.50}        \\
\textit{DeepSeek-R1} (ours)                      & 32B                  & 1.3K               & \textbf{39.00}        \\
\textit{Gemini-2.5-Flash} (ours)                 & 5B                   & 1.3K               & \textbf{45.50}        \\
\textit{Gemini-2.5-Pro} (ours)                   & -                    & 0.9K               & \textbf{51.00}        \\
\hline
\end{tabular}
}
\caption*{\textit{DeepSeek-R1} abbreviates \textit{DeepSeek-R1-Distill-Qwen}. (1)~\textbf{Top:} PnP baselines~\cite{rostami2025plug} with raw floating-point features; (2)~\textbf{Middle:} Raw features with a valid-options list appended; (3)~\textbf{Bottom (DiSC-AMC):} Discretized statistics, $5_\text{top}$, 5-way 1-shot classification.}
\end{table}

\subsection{Complexity Analysis}
\label{sec:complexity}

\subsubsection{Token Budget}
DiSC-AMC substantially reduces prompt length relative to PnP~\cite{rostami2025plug}, which requires $>$2{,}853 tokens per query. Two mechanisms drive this reduction: (i)~feature discretization compresses each continuous statistic from a multi-digit floating-point string to a single symbolic token, shrinking the feature block by $3$--$5{\times}$; and (ii)~DINOv2 candidate retrieval prunes the label space to $k_\text{top}{\le}5$ classes and correspondingly reduces the exemplar set. Together, these yield prompts of 785--1{,}315 tokens (Fig.~\ref{fig:ktops}), a reduction of over 50\%. The fine-tuned DiSC-AMC configuration (Table~\ref{tab:efficiency}) achieves this with prompts as short as 0.5K tokens while reaching 71.9\% accuracy on the same benchmark where PnP requires 2.9K tokens for 69.92\% (200B model).

\subsubsection{Parameter Budget}
In the fine-tuned setting, DiSC-AMC with a 7B model achieves 83.00\%, surpassing DenoMAE2.0 (82.40\%, 86M parameters) and all prior LLM-based baselines (69.92\% at 200B). In the training-free setting, our 5B \textit{Gemini-2.5-Flash} reaches 45.5\%, comparable to the 32B DeepSeek PnP baseline (47.8\%) at $84\%$ fewer parameters. These results show that careful prompt construction, not raw model scale, is the dominant factor in LLM-based AMC performance. The framework's modular design (swappable retrieval and encoder components, optional ``unknown'' class for open-set recognition) further supports deployment on resource-constrained hardware.

\section{Conclusion}
We introduced DiSC-AMC, a framework that reformulates automatic modulation classification as an LLM reasoning task through feature discretization, dynamic context pruning, and self-supervised candidate retrieval. A fine-tuned 7B-parameter LLM achieves 83.00\% in-distribution accuracy and 82.50\% OOD accuracy on our synthetic benchmark, surpassing all supervised baselines including DenoMAE2.0 (82.40\%). On the challenging RadioML.2018.01a dataset (24 classes), DiSC-AMC amplifies encoder backbones well beyond their standalone performance. With a fine-tuned DINOv2 backbone, DiSC-AMC reaches 76.25\% at $+20$\,dB compared to 9.58\% for the encoder alone; even with a frozen encoder, DiSC-AMC achieves 45.42\% versus 7.08\%. Systematic ablations confirm that FAISS-based retrieval outperforms centroid selection by over 30 percentage points, self-supervised embeddings consistently outperform statistical features, and coarse discretization (5 bins) aligns best with LLM reasoning. In the training-free setting, DiSC-AMC cuts prompt length by over 50\% and enables a 5B model to match a 32B baseline. A fundamental limitation remains: at extreme noise levels ($-30$\,dB), constellation diagrams become featureless, causing the encoder representations to collapse and reducing classification to random chance.

\balance
\bibliographystyle{IEEEtran}
\bibliography{egbib}

@article{oquab2023dinov2,
  title={Dinov2: Learning robust visual features without supervision},
  author={Oquab, Maxime and Darcet, Timoth{\'e}e and Moutakanni, Th{\'e}o and Vo, Huy and Szafraniec, Marc and Khalidov, Vasil and Fernandez, Pierre and Haziza, Daniel and Massa, Francisco and El-Nouby, Alaaeldin and others},
  journal={arXiv preprint arXiv:2304.07193},
  year={2023}
}

@inproceedings{faysal2024nmformer,
  title={Nmformer: A transformer for noisy modulation classification in wireless communication},
  author={Faysal, Atik and Rostami, Mohammad and Roshan, Reihaneh Gh and Wang, Huaxia and Muralidhar, Nikhil},
  booktitle={2024 33rd Wireless and Optical Communications Conference (WOCC)},
  pages={103--108},
  year={2024},
  organization={IEEE}
}

@article{faysal2025denomae,
  title={{D}enomae: {A} multimodal autoencoder for denoising modulation signals},
  author={Faysal, Atik and Boushine, Taha and Rostami, Mohammad and Roshan, Reihaneh Gh and Wang, Huaxia and Muralidhar, Nikhil and Sahoo, Avimanyu and Yao, Yu-Dong},
  journal={IEEE Communications Letters},
  year={2025},
  publisher={IEEE}
}

@INPROCEEDINGS{rostami2025plug,
  author={Rostami, Mohammad and Faysal, Atik and Roshan, Reihaneh Gh. and Wang, Huaxia and Muralidhar, Nikhil and Yao, Yu-Dong},
  booktitle={2025 IEEE 34th Wireless and Optical Communications Conference (WOCC)}, 
  title={Plug-and-{P}lay {AMC}: {C}ontext {I}s {K}ing in {T}raining-Free, {O}pen-{S}et {M}odulation with {LLM}s}, 
  year={2025},
  volume={},
  number={},
  pages={345-350},
  doi={10.1109/WOCC63563.2025.11082201}}

@article{faysal2025denomae2,
  title={Deno{MAE}2. 0: {I}mproving {D}enoising {M}asked {A}utoencoders by {C}lassifying {L}ocal {P}atches},
  author={Faysal, Atik and Rostami, Mohammad and Boushine, Taha and Roshan, Reihaneh Gh and Wang, Huaxia and Muralidhar, Nikhil},
  journal={arXiv preprint arXiv:2502.18202},
  year={2025}
}

@article{dobre2007survey,
  title={{S}urvey of automatic modulation classification techniques: classical approaches and new trends},
  author={Dobre, Octavia A and Abdi, Ali and Bar-Ness, Yeheskel and Su, Wei},
  journal={IET communications},
  volume={1},
  number={2},
  pages={137--156},
  year={2007},
  publisher={IET}
}

@article{peng2018modulation,
  title={{M}odulation classification based on signal constellation diagrams and deep learning},
  author={Peng, Shengliang and Jiang, Hanyu and Wang, Huaxia and Alwageed, Hathal and Zhou, Yu and Sebdani, Marjan Mazrouei and Yao, Yu-Dong},
  journal={IEEE Transactions on Neural Networks and Learning Systems},
  volume={30},
  number={3},
  pages={718--727},
  year={2018},
  publisher={IEEE}
}

@ARTICLE{8963964,
  author={Huynh-The, Thien and Hua, Cam-Hao and Pham, Quoc-Viet and Kim, Dong-Seong},
  journal={IEEE Communications Letters}, 
  title={{MCNet}: {A}n {E}fficient {CNN} {A}rchitecture for {R}obust {A}utomatic {M}odulation {C}lassification}, 
  year={2020},
  volume={24},
  number={4},
  pages={811-815},
  doi={10.1109/LCOMM.2020.2968030}}

@article{jassim2022comparison,
  title={{C}omparison of {A}utomatic {M}odulation {C}lassification {T}echniques.},
  author={Jassim, Salah Ayad and Khider, Ibrahim},
  journal={J. Commun.},
  volume={17},
  number={7},
  pages={574--580},
  year={2022}
}

@inproceedings{fontaine2024towards,
  title={Towards a wireless physical-layer foundation model: Challenges and strategies},
  author={Fontaine, Jaron and Shahid, Adnan and De Poorter, Eli},
  booktitle={2024 IEEE International Conference on Communications Workshops (ICC Workshops)},
  pages={1--7},
  year={2024},
  organization={IEEE}
}

@inproceedings{mirarab2007robust,
  title={{R}obust modulation classification for PSK/QAM/ASK using higher-order cumulants},
  author={Mirarab, MR and Sobhani, MA},
  booktitle={2007 6th International Conference on Information, Communications \& Signal Processing},
  pages={1--4},
  year={2007},
  organization={IEEE}
}

@inproceedings{10.1145/882082.882086,
author = {Lin, Jessica and Keogh, Eamonn and Lonardi, Stefano and Chiu, Bill},
title = {{A} symbolic representation of time series, with implications for streaming algorithms},
year = {2003},
isbn = {9781450374224},
publisher = {Association for Computing Machinery},
address = {New York, NY, USA},
url = {https://doi.org/10.1145/882082.882086},
doi = {10.1145/882082.882086},
booktitle = {Proceedings of the 8th ACM SIGMOD Workshop on Research Issues in Data Mining and Knowledge Discovery},
pages = {2–11},
numpages = {10},
location = {San Diego, California},
series = {DMKD '03}
}

@misc{dong2024surveyincontextlearning,
      title={{A} {S}urvey on {I}n-context {L}earning}, 
      author={Qingxiu Dong and Lei Li and Damai Dai and Ce Zheng and Jingyuan Ma and Rui Li and Heming Xia and Jingjing Xu and Zhiyong Wu and Tianyu Liu and Baobao Chang and Xu Sun and Lei Li and Zhifang Sui},
      year={2024},
      eprint={2301.00234},
      archivePrefix={arXiv},
      primaryClass={cs.CL},
      url={https://arxiv.org/abs/2301.00234}, 
}

@inproceedings{liu-etal-2022-makes,
    title = "{W}hat {M}akes {G}ood {I}n-{C}ontext {E}xamples for {GPT}-3?",
    author = "Liu, Jiachang  and
      Shen, Dinghan  and
      Zhang, Yizhe  and
      Dolan, Bill  and
      Carin, Lawrence  and
      Chen, Weizhu",
    editor = "Agirre, Eneko  and
      Apidianaki, Marianna  and
      Vuli{\'c}, Ivan",
    booktitle = "{P}roceedings of {D}eep {L}earning {I}nside {O}ut ({DeeLIO} 2022): The 3rd Workshop on Knowledge Extraction and Integration for Deep Learning Architectures",
    month = may,
    year = "2022",
    address = "Dublin, Ireland and Online",
    publisher = "Association for Computational Linguistics",
    url = "https://aclanthology.org/2022.deelio-1.10/",
    doi = "10.18653/v1/2022.deelio-1.10",
    pages = "100--114"
}

@misc{yang2024autoiclincontextlearninghuman,
      title={{Auto-ICL}: {In-Context Learning without Human Supervision}}, 
      author={Jinghan Yang and Shuming Ma and Furu Wei},
      year={2024},
      eprint={2311.09263},
      archivePrefix={arXiv},
      primaryClass={cs.LG},
      url={https://arxiv.org/abs/2311.09263}, 
}

@article{comanici2025gemini,
  title={{Gemini 2.5: Pushing the frontier with advanced reasoning, multimodality, long context, and next generation agentic capabilities}},
  author={Comanici, Gheorghe and Bieber, Eric and Schaekermann, Mike and Pasupat, Ice and Sachdeva, Noveen and Dhillon, Inderjit and Blistein, Marcel and Ram, Ori and Zhang, Dan and Rosen, Evan and others},
  journal={arXiv preprint arXiv:2507.06261},
  year={2025}
}

@article{douze2024faiss,
  title={The faiss library},
  author={Douze, Matthijs and Guzhva, Alexandr and Deng, Chengqi and Johnson, Jeff and Szilvasy, Gergely and Mazar{\'e}, Pierre-Emmanuel and Lomeli, Maria and Hosseini, Lucas and J{\'e}gou, Herv{\'e}},
  journal={IEEE Transactions on Big Data},
  year={2025},
  publisher={IEEE}
}

@article{ansari2025attention,
  title={Attention-enhanced hybrid automatic modulation classification for advanced wireless communication systems: A deep learning-transformer framework},
  author={Ansari, Sam and Alnajjar, Khawla A and Majzoub, Sohaib and Almajali, Eqab and Jarndal, Anwar and Bonny, Talal and Hussain, Abir and Mahmoud, Soliman},
  journal={IEEE Access},
  year={2025},
  publisher={IEEE}
}

@ARTICLE{8267032,
  author={O’Shea, Timothy James and Roy, Tamoghna and Clancy, T. Charles},
  journal={IEEE Journal of Selected Topics in Signal Processing}, 
  title={Over-the-Air Deep Learning Based Radio Signal Classification}, 
  year={2018},
  volume={12},
  number={1},
  pages={168-179},
  doi={10.1109/JSTSP.2018.2797022}}

@article{guo2025deepseek,
  title={Deepseek-r1: Incentivizing reasoning capability in llms via reinforcement learning},
  author={Guo, Daya and Yang, Dejian and Zhang, Haowei and Song, Junxiao and Wang, Peiyi and Zhu, Qihao and Xu, Runxin and Zhang, Ruoyu and Ma, Shirong and Bi, Xiao and others},
  journal={arXiv preprint arXiv:2501.12948},
  year={2025}
}

@article{zhang2022deep,
  title={Deep learning based automatic modulation recognition: Models, datasets, and challenges},
  author={Zhang, Fuxin and Luo, Chunbo and Xu, Jialang and Luo, Yang and Zheng, Fu-Chun},
  journal={Digital Signal Processing},
  volume={129},
  pages={103650},
  year={2022},
  publisher={Elsevier}
}

@article{purohit2025sample,
  title={Sample efficient demonstration selection for in-context learning},
  author={Purohit, Kiran and Venktesh, V and Bhattacharya, Sourangshu and Anand, Avishek},
  journal={arXiv preprint arXiv:2506.08607},
  year={2025}
}

@article{bhope2025optiseq,
  title={Optimizing Example Ordering for In-Context Learning},
  author={Bhope, Rahul Atul and Venkateswaran, Praveen and Jayaram, K. R. and Isahagian, Vatche and Muthusamy, Vinod and Venkatasubramanian, Nalini},
  journal={arXiv preprint arXiv:2501.15030},
  year={2025}
}

@inproceedings{zhang2024large,
  title={Large language models in wireless application design: In-context learning-enhanced automatic network intrusion detection},
  author={Zhang, Han and Sediq, Akram Bin and Afana, Ali and Erol-Kantarci, Melike},
  booktitle={GLOBECOM 2024-2024 IEEE Global Communications Conference},
  pages={2479--2484},
  year={2024},
  organization={IEEE}
}

@inproceedings{guo2024demo,
  title={What Makes a Good Order of Examples in In-Context Learning},
  author={Guo, Qi and Wang, Leiyu and Wang, Yidong and Ye, Wei and Zhang, Shikun},
  booktitle={Findings of the Association for Computational Linguistics: ACL 2024},
  pages={14892--14904},
  year={2024}
}

@article{yin2024deeper,
  title={Deeper Insights Without Updates: The Power of In-Context Learning Over Fine-Tuning},
  author={Yin, Qingyu and He, Xuzheng and Deng, Luoao and Leong, Chak Tou and Wang, Fan and Yan, Yanzhao and Shen, Xiaoyu and Zhang, Qiang},
  journal={arXiv preprint arXiv:2410.04691},
  year={2024}
}

@article{hong2025glm,
  title={Glm-4.5 v and glm-4.1 v-thinking: Towards versatile multimodal reasoning with scalable reinforcement learning},
  author={Hong, Wenyi and Yu, Wenmeng and Gu, Xiaotao and Wang, Guo and Gan, Guobing and Tang, Haomiao and Cheng, Jiale and Qi, Ji and Ji, Junhui and Pan, Lihang and others},
  journal={arXiv preprint arXiv:2507.01006},
  year={2025}
}

\end{document}